# Analysis of instruction-based LLMs' capabilities to score and judge text-input problems in an academic setting

Valeria Ramirez-Garcia, David de-Fitero-Dominguez, Antonio Garcia-Cabot, Eva Garcia-Lopez

Universidad de Alcalá

**ABSTRACT**

Large language models (LLMs) offer an opportunity to act as evaluators. This capacity has been studied by methods like LLM-as-a-Judge and fine-tuned judging LLMs. In the field of education, LLMs have been studied as assistant tools for students and teachers. Our research investigates the use of LLM-driven automatic evaluation systems for academic Text-Input Problems through rubrics; with the objective of bringing useful, automated and customized feedback to students. To do this, we propose five evaluation systems that have been tested on a custom dataset of 110 answers about computer science from higher education students. The first one employs JudgeLM, while the rest have been tested with using both Llama-3.1-8B and DeepSeek-R1-Distill-Llama-8B. The evaluation systems are the following: The JudgeLM evaluation, which uses the model's single answer prompt to obtain a score; Reference Aided Evaluation, which uses a correct answer as a guide aside from the original context of the question; No Reference Evaluation, which only uses the context to generate an evaluation; Additive Evaluation, which uses the addition of atomic criteria; and Adaptive Evaluation, which is an evaluation done with generated criteria fitted to each question. All the explained evaluation methods have been compared with the results of a human evaluator. The conclusions of the research determine that the best method to automatically evaluate and score Text-Input Problems using LLMs is Reference Aided Evaluation. With the lowest median absolute deviation of 0.945 and the lowest root mean square deviation of 1.214 when compared to human evaluation, Reference Aided Evaluation offers fair scoring as well as insightful and complete evaluations. Other methods such as Additive and Adaptive Evaluation fail to provide good results in short, concise answers. No Reference Evaluation lacks information needed to correctly assess questions, demonstrating that the inclusion of a reference answer is vital for correct evaluations. Lastly, JudgeLM Evaluations have not provided good results due to limitations of the model's context length and its lack of adaptability to transform the problem of judging LLMs into judging students' answers. As a result, we conclude that Artificial Intelligence-driven automatic evaluation systems, aided with proper methodologies such as Reference Aided Evaluation, show potential to work as complementary tools to other academic resources, improving learning experience for all in the academic field.

# 1 INTRODUCTION

Generative Artificial Intelligence (AI) tools, and particularly instruction-based chat assistants, have become widely used in the academic field. Discussions have arisen about ethical concerns on its use from students (Wyk, 2024), but these tools can also have a positive impact in the academic setting (J. Wang et al., 2024). Leveraging Large Language Models' (LLM) capabilities and human input from educational professionals, we aim to bring personalized feedback to support students in their fields of study. The focus of our study is the evaluation of Text-Input Problems (TIP), which offer unique insight into key aspects of education goals such as knowledge, comprehension, application, analysis, synthesis and evaluation (Brookhart, 2013). Automatic evaluation often relies on easily gradable questions such as Multiple-Choice Questions or numeric entry questions. The evaluation of Text-Input Problems is a unique challenge, since there is not one predetermined right answer, and evaluation criteria are more complex.

In this paper, we propose the use of LLMs as evaluators for TIP to gain deeper insight into the capabilities of LLMs to identify and judge complex criteria. To study this, first, a questionnaire has been generated to obtain answers for TIP from real students from different academic levels. Then, evaluations of these answers are generated with three models: one fine-tuned for judging, JudgeLM (Zhu et al., 2023), and two instruction-based models without further training, Llama3.1 and DeepSeek-R1-Distill-Llama.

The rest of the paper is as follows: In section 2, we present the state of the art when it comes to LLM evaluation in academic settings. Section 3 explains our methodology in the application models to evaluate academic goals. Section 4 and Section 5 present and discuss the results obtained, respectively, and finally, Section 6 draws the conclusions of the study and presents the limitations of this study.

## 2 STATE OF THE ART

The integration of Large Language Models in education, particularly for validating and correcting student answers, represents a promising development (Geschwind et al., 2024; Meyer et al., 2024). However, to be truly effective, these systems must be context-aware (Baral et al., 2024), capable of providing feedback (Kinder et al., 2025), and tailored to the pedagogical intent of each task (Steinert et al., 2024).

One of the key motivations of AI systems is automatization. LLMs bring a special opportunity to automatize processes where traditionally human intervention was strictly necessary (Chiang & Lee, 2023). The use of LLMs to perform validations has been extensively studied. In the context of teaching and learning, automating the correction and evaluation of student work is particularly valuable. Perfecting the process of reviewing answers would allow for faster development and quicker results (Kim et al., 2024; Zhu et al., 2023).

AI, and particularly Generative AI, have already been used in the field of education, especially in gamification (Abbes et al., 2024; Šego & Gakić, 2024). It has shown to improve motivation, and it focuses on customization, by analyzing the learning profile of the student to best adapt to their needs. However, it still has challenges to overcome such as adherence to ethical standards. The role of this technology in education has advantages both for teaching and learning (Jahić et al., 2024; Kumar et al., 2024). In teaching, AI can help to plan lessons, generate content, and administrate courses. In learning, the main appeal of these tools is to offer intelligent tutoring and interactive learning experience.

One key feature of intelligent tutoring is the ability to interact through questions and receive personalized feedback. For the interactive experience, on-demand content for study support is of vital importance. For this, Automated Question Generation (AQG) has proved to be useful. Question generation (Mulla & Gharpure, 2023) for education has three main purposes: knowledge acquisition, knowledge assessment and tutorial dialogues (Le et al., 2014). Most of the research in question generation is concerned with knowledge assessment of domain-specific free response questions (Kurdi et al., 2020). This matches the purposes of this study.

Through AQG, LLMs turn into a powerful academic assistant tool. However, from the students' perspective, questions are not enough to guarantee a quality interactive learning experience. This means that for AQG systems to be fully integrated in learning environments, they must be accompanied by feedback. Feedback is one of the most powerful tools for effective learning (Hattie & Timperley, 2007), be it grading or providing explanations. For these questions to be part of a complete system, they also need to be graded. On the other hand, feedback through AI is given mainly by the following three types of AI-powered systems: Automatic Grading/Review Systems, Automatic Feedback Systems and LLMs-as-a-judge.

Automatic Grading Systems have already been applied to the academic field (Flodén, 2025; Tyser et al., 2024; J. Wang et al., 2024). The main strengths of these automated systems are replicability, scalability and the possibility of eliminating biases. Since not all academic resources can be graded by humans, due to the number of students, Automatic Grading Systems can offer consistent evaluation even at large scale, which is critical in online or large-group settings. Other kinds of automatic evaluation systems have been used in other areas, such as interviews (Uppalapati et al., 2025).

Academic Feedback is not only represented through grades, but with an analysis of performance as well. Automatic feedback systems aim to provide meaningful and personalized evaluations. These systems have been traditionally implemented with correction rules in the field of computer science for introductory exercises (Singh et al., 2013). Applications for this technology in academic fields have also been recently studied with LLMs to generate feedback for academic papers (Tyser et al., 2024) and as an assistant tool for academic work (J. Wang et al., 2024). The benefits of automated feedback have also been tested, showing that analysis must also be accompanied with scoring in order to improve students' self-assessment accuracy (Liebenow et al., 2025), showing both components are key aspects of presenting useful evaluations.

The last technique for evaluating performance is the use of LLM-as-judges. Although the main objective of this approach is to judge the performance of LLMs themselves (Kim et al., 2024; Zheng et al., 2023; Zhu et al., 2023), the techniques applied allow us to analyze LLM capabilities when it comes to working with diverse rubrics. This opens up new possibilities for evaluating responses through human-aligned rubrics. This explores the possibility to judge the performance of LLMs beyond traditional metrics to better suit human preference and evaluate complex scenarios such as multiturn chats. Finetuning models to this task (as done by Kim et al., 2024; Zhu et al., 2023) allows us to overcome known biases that affect the resulting evaluations, such as knowledge bias or position bias (Zhu et al., 2023). Techniques such as relying on a reference answer to evaluate answers are methods which have been used in LLM-as-judges (Kim et al., 2024) and which fit our purposes for academic evaluations.

In education, different types of questions and exercises have different purposes and must be evaluated accordingly (Brookhart, 2013). With proper instructions, LLMs can approximate human evaluations (Chiang & Lee, 2023), allowing for rich feedback relevant not only to the student's answer, but also to the specific intent of the exercise itself. Including relevant criteria is a key aspect for fair academic evaluation. The inclusion of AI systems and agents in online learning has been studied (Viswanathan et al., 2022) and shows promise to increase flexibility and customization for the learner.

While previous work has applied LLMs to essays and exams, LLMs still show inconsistent and inaccurate results in the context of academic evaluations (Seo et al., 2025). This study focuses specifically on correcting student answers to open-ended text-input questions. The work in this field, also called Automatic Short Answer Grading (ASAG), has advanced in recent years. New techniques such as dividing answers into lexical and semantical components have been applied to obtain scores finding success in achieving human-like scoring (del Gobbo et al., 2023).

Evaluating text-input questions offers a unique challenge where evaluation criteria and prompts are of key importance. LLMs still show inconsistent and inaccurate results in the context of academic evaluations (Seo et al., 2025). This paper explores different techniques based on Automatic Review Systems, Automatic Feedback Systems and LLM-as-judges, leveraging educational context embedded in the prompt to enhance the accuracy and relevance of AI-generated feedback and how the evaluation criteria can help guide evaluations. The aim is not only to generate accurate scoring, but also to offer feedback and the rationale behind the evaluation to increase the interpretability of the results.

## 3 METHODOLOGY

In this section, the methodology for each evaluation process is presented, as well as every material used, including the models, questions and prompts. Additionally, for each method, explanations are included to further contextualize the objective for each of the methodologies presented.

### 3.1 Evaluation Process

To test the capabilities of different models for our purpose of generating fair evaluations and meaningful feedback, in addition to the human evaluation, five automated scoring pipelines have been proposed, as shown in Figure 1.

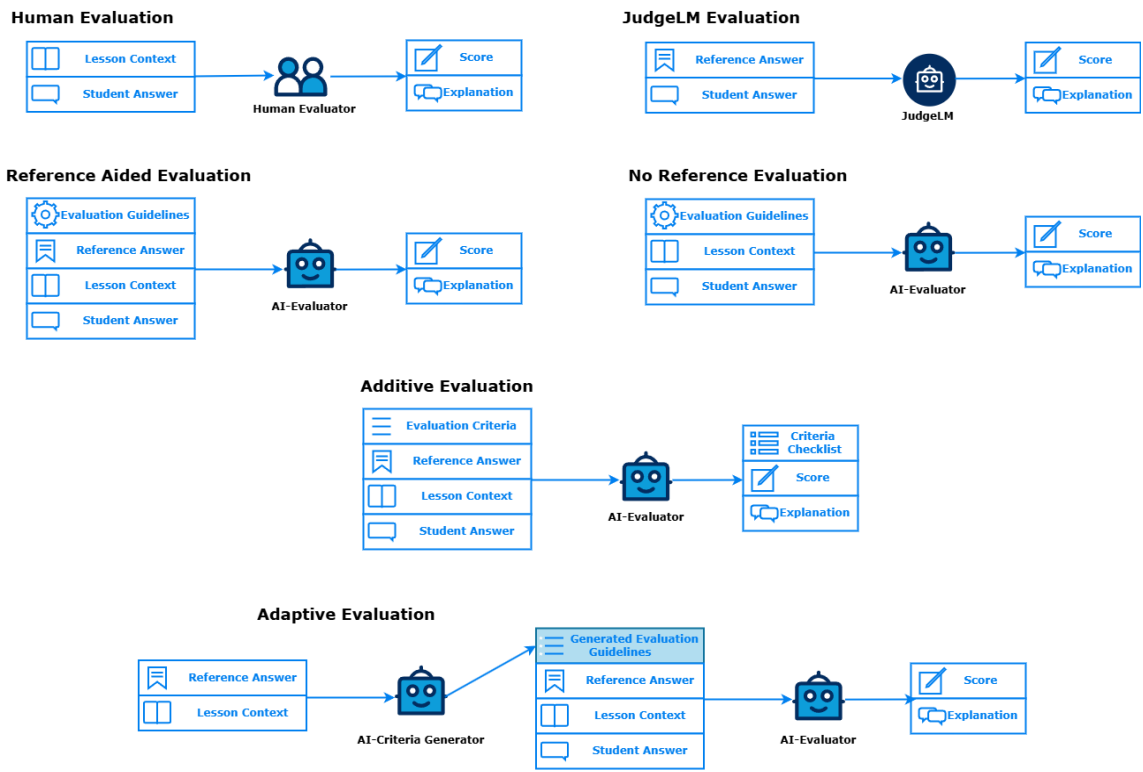

**Figure 1. Process of evaluation for each method.**

The questions used to gather the questions dataset have been generated from the MIT's Massive Online Open Course (MOOC) "Introduction to Computer Science and Programming in Python", with help of a Llama-3.1-8B model and hand-selected by the evaluator. Each question was generated with its corresponding reference answer, which has also been manually checked by the evaluator. With the final set of 10 questions about Python and Computer Science, we obtained 110 unique answers from 11 students at Universidad de Alcalá from engineering related fields in higher education: 4 Undergraduate students, 3 Master's students and 4 PhD students, with ages ranging 19-36. The questions used can be seen in Table 1.

Table 1. Questions used for gathering the answers' dataset.

| 1 | What is the main idea behind the complexity of an algorithm? |
|---|---|
| 2 | What is the main idea behind treating debugging as a search problem? |
| 3 | What is the purpose of using the 'raise' keyword in Python? |
| 4 | What is the main goal of a search algorithm? |
| 5 | What are the two fundamental things that a computer can do? |
| 6 | What is a string in Python? |
| 7 | What is the primary characteristic of an algorithm that exhibits quadratic complexity? |
| 8 | What is the complexity of a function h(n) that adds up the digits of a number together? |
| 9 | What is the main idea behind object-oriented programming? |
| 10 | What is the main goal of analyzing algorithms to determine their complexity? |

Our point of reference is data obtained through human evaluation. The evaluator has assigned scores based on both the reference answer for each question and the base material of the course. Human evaluation may also contain biases, because personal preferences and criteria affect academic judgement (Gil-Hernández et al., 2024). Due to this, the objective is not for models to replicate the scoring results of human evaluation. This data serves as a comparison to judge fairness and analyze and compare explicability between models.

Common to all evaluation pipelines are the obtained score and explanation. All scoring methods work on a 0-4 Likert scale, 0 being the lowest score possible and 4 being the highest. Explanations are complementary to the score and must explain the reasoning behind it (e.g., what's missing from the answer or what criteria were followed to reward the score). Section 3.3 Evaluation Methodologies goes into further detail about each method.

## 3.2 Models Used for Evaluation

In the following sections, the models that have been used are presented alongside the chosen characteristics like temperature and model size. The three employed models for this study are JudgeLM, as a fine-tuned instruction-based judge, and Llama and DeepSeek as pretrained instruction-based models. With the objective of making this technology accessible, all the chosen models are open-source small scale models. The selected temperature was 0.2 for all models to have greater consistency in responses.

Llama and DeepSeek were selected as representative state-of-the-art instruction-tuned language models due to their strong performance across a range of benchmark tasks (DeepSeek-AI et al., 2025; Grattafiori et al., 2024). These models were also utilized to assess the potential impact of Chain-of-Thought (CoT) fine-tuning on performance in the evaluated task by comparing the results of Llama and DeepSeek-R1, the latter being pretrained with this specific approach (CoT) (DeepSeek-AI et al., 2025). Additionally, JudgeLM was chosen as the evaluation model, given its fine-tuning for use as an LLM-as-a-judge system. Its training specifically aims to mitigate biases in model-based evaluation, making it an appropriate and reliable tool for comparative assessment.

With the exception of JudgeLM, which was used with its own specialized judge-based approach, all other evaluation methodologies were tested with both Llama and DeepSeek.

### 3.2.1 JudgeLM

To assess LLM's biases when judging (P. Wang et al., 2023), JudgeLM has been chosen as a judge. JudgeLM is a model fine-tuned to evaluate LLM and chatbot performance. Whilst this does not exactly match our purpose, JudgeLM has been trained to score and judge answers while avoiding biases like knowledge bias or position bias (Zhu et al., 2023), making it a good candidate for our academic evaluation.

This model has a unique pipeline based on single answer judging, taking a pair of answers where one is the reference answer (which is assigned the maximum score) and the other is the student's answer for which we need the evaluation. The prompt for this (Annex A) has been adapted from the examples for single answer judging presented for the model (Zhu et al., 2023).

The biggest drawback with the utilization of this model is its limited token size, which is only 2048 tokens. This means that the context for the evaluation is limited only to the reference answer, since no further content can be added to aid in evaluation. This poses a great problem for academic evaluations, since contextualization and base material to pull from are essential for a proper evaluation (Seo et al., 2025).

The version of the model chosen for this experiment is JudgeLM-7B-v1.0[1].

### 3.2.2 Llama3.1

Llama3.1 is a flexible instruction model. Instruction-based models offer us the adaptability necessary for understanding the grading task without further training and fine-tuning. For this, four evaluation pipelines are tested for Llama3.1: Reference Aided Evaluation, No Reference Evaluation, Additive Evaluation and Adaptive Evaluation. This variety of methods illustrates possible strengths and weaknesses when generating evaluations for TIP.

The chosen version is Llama-3.1-8B-Instruct[2], because smaller models require less investment in resources for their deployment.

### 3.2.3 DeepSeek-R1

DeepSeek-R1's main appeal is its reasoning capabilities (DeepSeek-AI et al., 2025). Explicability and reasoning are of great importance when it comes to generating evaluations and feedback: scoring should be thought out and justified to be fair. The same four evaluation pipelines as Llama3.1's have been tested for DeepSeek-R1.

The chosen version of the model is a distilled trained version of a smaller Llama model: DeepSeek-R1-Distill-Llama-8B[3], chosen because of its reduced size and to act as a point of comparison on a base Llama-8B model.

## 3.3 Evaluation Methodologies

In this section, we explore all the different evaluation strategies tested with Llama3.1 and DeepSeek-R1. The objective of these methodologies is to test different approaches to generate evaluations, thus helping us to detect potential strengths and weaknesses that LLMs present in the context of academic evaluations. LLMs excel in understanding language but can struggle when presented with domain-specific problems (Chen et al., 2024; Yang et al., 2023), which is a clear weakness when applying these tools. All methods have been tested with a particular motivation that is explained in each following subsection.

### 3.3.1 Reference Aided Evaluation

Text-input problems are harder to evaluate than single-answer questions. To mitigate this, the inclusion of a reference answer in Reference Aided Evaluation aims to guide the model on what can be expected of a good answer to the question. This not only helps to specify what content the student's answer should contain, but also helps the model further understand the intent of the question. LLMs excel in answering general questions, not factual knowledge (Chiang & Lee, 2023; Flodén, 2025). Therefore, the addition of a reference answer allows for different answers to be recognized as correct if they follow the key points illustrated in the reference answer, allowing for fair evaluations in the face of varied answers (Kim et al., 2024). This introduces one problem: we wish for the evaluation to be flexible as well as correct. An answer can be right while barely matching with the provided reference. Results of this method can help illustrate if the inclusion of this reference helps or harms the evaluation.

---
[1] https://huggingface.co/BAAI/JudgeLM-7B-v1.0
[2] https://huggingface.co/meta-llama/Llama-3.1-8B-Instruct
[3] https://huggingface.co/deepseek-ai/DeepSeek-R1-Distill-Llama-8B

Prompts for this approach can be seen in Annex B. The input for the evaluation includes the evaluation guidelines, the lesson content, a reference answer and the student answer to be evaluated, as shown in Figure 1. For each score in the evaluation guidelines, requirements get stricter: the answer must show increasing levels of depth in the student's understanding of the topic. For example, a score of 2 can be awarded by mentioning the key aspects of the answer to the question, whilst a score of 4 requires complementary explanations and examples to be awarded.

These evaluation guidelines are based in the work of Brookhart2013). They are non-specific, meaning that they are not adapted to the content being judged. Therefore, although this experiment works with a dataset of answers about computer science, these criteria could be applied to any other field.

### 3.3.2 No Reference Evaluation

No Reference Evaluation works mainly as a point of comparison with Reference Aided Evaluation. The process of generation is the same as in the previous approach but omitting the reference answer. This allows the model to evaluate based only on the lesson's content.

Whilst Reference Aided Evaluation had the problem of "enforcing" certain structure or content in the expected answer, No Reference Evaluation might lack tools to correctly evaluate an answer. Text-input problems can ask the student to apply their gained knowledge in ways not explicitly specified by the teaching material. For example, in the resolution of problems, classification, analysis of new but related concepts or the creative output of new ideas related to the topic.

Prompts for this approach can be seen in

**Figure 15. User prompt for Reference Aided Evaluation and Adaptive Evaluation.**

Annex C. Instructions and criteria are equal to those used in Reference Aided Evaluation.

### 3.3.3 Additive Evaluation

Additive Evaluation consists of defining a series of atomic criteria, each of which score a certain number of points, totaling to a maximum amount of 4. Each criterion can award a different number of points depending on their importance. To the best of our knowledge, this represents the first application of this boolean criteria approach to the purpose of academic evaluation with LLMs.

While hypothetically this approach would allow for further customization, adapting it to non-specific criteria undermines this advantage. Additive criterion are usually applied to long and complex evaluation tools, such as essays (Brookhart, 2013). This approach could be advantageous to define a set of human-made criteria to judge longer pieces of writing. For our purposes, however, defining criteria for each question is inconvenient, as, ideally, this system would work with a large range of questions.

Nonetheless, this approach was tested to analyze the capabilities of LLMs to correctly identify and judge based on these atomic criteria, though on a smaller scale. Prompts for this method can be seen in Annex D.

### 3.3.4 Adaptive Evaluation

Adaptive Evaluation is a two-step approach to the evaluation generation process. As mentioned earlier, evaluation can benefit from personalized criteria to create more meaningful evaluations and feedback.

Step one consists in generating question-specific criteria for each question. Prompts of Annex E have been used for this generation. They include instructions to create criteria going from 0-4, similar to what was done for Reference Aided Evaluation and No Reference Evaluation. Criteria must be specific to the question and use the lesson material as support. Step two is the evaluation itself, and its format is the same as the one used in Reference Aided Evaluation, where the criteria are filled with the generated criteria for each question.

However, this leaves the important task of defining what exactly is relevant to include in an answer to the model. To avoid the creation of unrelated or unwanted criteria, the reference answer acts as a support to guide the generation. This can ensure that the model's thought process for the evaluation guidelines is relevant to the scenario.

## 4 RESULTS

Results of scoring for each method are shown in both a heatmap, with values divided by question, and a histogram, alongside with statistical measures. For additive evaluation, complementary tables have been added to show percentage for criteria fulfillment. Examples of complete evaluations with scores and explanations are included to discuss strengths and weaknesses of the explainability and reasoning of the tested methods.

Table 2. Score statistics for each evaluation method.

| Model | Evaluation | N | Mean | Median | Std |
|---|---|---|---|---|---|
| **None** | **Human** | 110 | 2.636 | 3.0 | 1.171 |
| **JudgeLM** | **JudgeLM** | 73 | 2.822 | 3.0 | 0.653 |
| **Llama-3.1** | **Reference Aided** | 110 | 1.818 | 2.0 | 0.940 |
| | **No Reference** | 110 | 1.455 | 1.0 | 0.874 |
| | **Additive** | 110 | 1.009 | 1.0 | 0.439 |
| | **Adaptive** | 110 | 1.382 | 1.0 | 0.835 |
| **DeepSeek-R1** | **Reference Aided** | 110 | 1.091 | 1.0 | 0.873 |
| | **No Reference** | 110 | 0.855 | 1.0 | 0.811 |
| | **Additive** | 110 | 0.218 | 0.0 | 0.882 |
| | **Adaptive** | 110 | 0.845 | 1.0 | 0.609 |

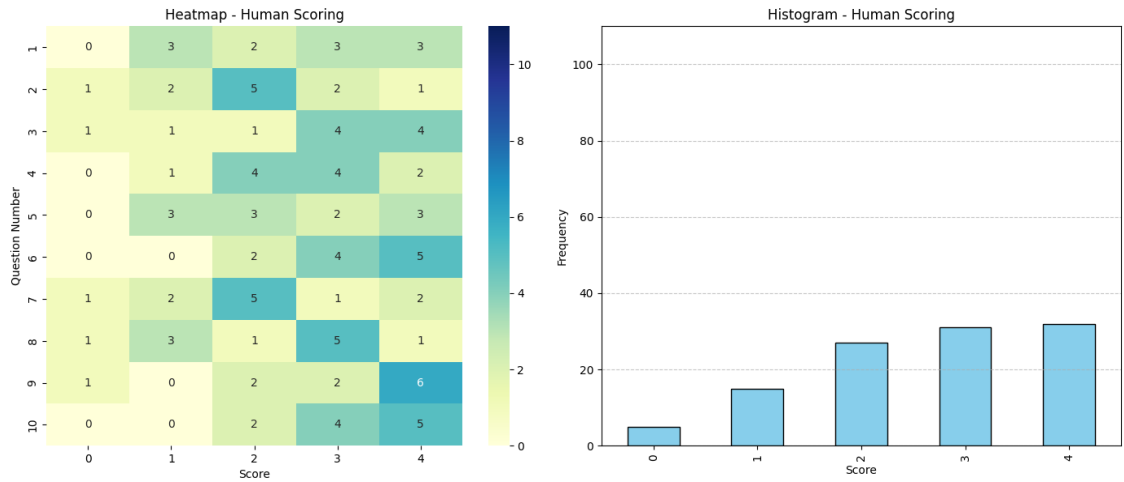

**Figure 2. Heatmap and histogram of scoring results obtained through human evaluation.**

The results of human evaluation (Figure 2) are our baseline. This figure shows the number of answers per question that got that score on the left and, on the right, the total number of answers for each score. While the histogram offers a general comparison of the distribution of scores, the heatmap helps illustrate whether the models agree or not with the human evaluator's judgement.

When talking about the statistical results of all other methods, it will be in reference to human evaluation. When evaluated by a human, most answers get a passing grade, as indicated by a mean of 2.636 and, compared with the rest of the methods, human evaluation has the highest standard deviation (1.171) (Table 2). This suggests that evaluators perceive a greater diversity in the quality of the answers compared to the automatic evaluation pipelines.

## 4.1 JudgeLM Evaluation

The JudgeLM's prompt results are illustrated in Figure 3. A unique value of -1 has been introduced to indicate answers that could not be evaluated. This is due to the token limit, which is the first observable drawback of the approach with this model. Not only are we forced to omit the larger context and rely only on a reference answer that might offer limited information for contextualization, but this limit is also unfit to handling the length of many of the evaluated questions.

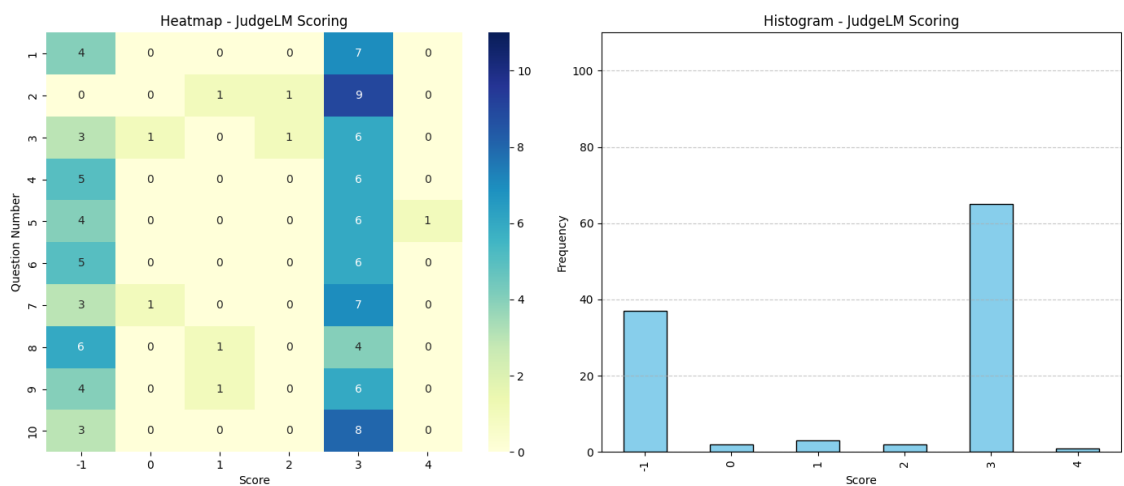

**Figure 3. Heatmap and histogram of scoring results obtained by JudgeLM.**

In 37 out of 110 answers, the length of the prompt, the reference answer and the student's answers combined exceeded the token limit. Therefore, for this case study, 33.63% of answers could not be evaluated. In a question format that incentivizes lengthy explanations, this proves to be an important obstacle to making use of this model for our purposes.

JudgeLM's results show poor capabilities to understand and grade answers. Out of the 73 evaluations, 65 of them have been rated with a score of 3, meaning that 89.04% of answers share the same score, giving it a standard deviation score of 0.653 (Table 2). While this is not the lowest standard deviation, being the third lowest, this is also affected by the lower number of examples available from JudgeLM's results. This means outliers have a bigger impact on increasing standard deviation in comparison to the rest of the models.

Table 3. JudgeLM evaluation example.

| JudgeLM Evaluation | |
|---|---|
| Question | What is the primary characteristic of an algorithm that exhibits quadratic complexity? |
| Reference Answer | The presence of nested loops, which can lead to a worst-case scenario where the algorithm performs n squared operations, where n is the size of the input. |
| Student | An algorithm that exhibits quadratic complexity means that for an input n the complexity scales quadratically, for a higher size the complexity is also higher. |
| Human Evaluator | Score: 1<br>Quadratic complexity does scale "quadratically" but this statement offers no additional information. For other complexities, like n, for a higher n there is also higher complexity: which means the answer is neither insightful or relevant to quadratic complexity specifically. |
| AI Evaluator (JudgeLM) | Score: 3<br>Both students provided correct answers, but neither of them mentioned the term 'quadratic complexity' which is a key term in the question. Therefore, their answers are not fully accurate. However, they both correctly identified that an algorithm with quadratic complexity would<br>perform n^2 operations, where n is the size of the input. Therefore, they both receive a score of 3 for providing relevant and somewhat accurate information, but lacking in detail and precision. |

Table 3 shows one example evaluation that illustrates JudgeLM's judgement. The model cannot fully comprehend what the question is asking and rewards an answer that repeats information from the question itself. While other approaches focus on content, JudgeLM still rewards clarity and proper vocabulary over factuality due to its fine-tuning, which prioritizes characteristics related to helpfulness, not necessarily correctness (Zhu et al., 2023).

Consistency in explanations is also an issue with evaluations. To evaluate single answers with JudgeLM, a specific prompt pattern is used, in which the reference answer is provided as part of a pair and given the maximum score (4), then the model continues from the assigned score of the first answer (the reference) to judge the second one. In this example, the model explains how the reference answer has a score of 3 when its assigned score is 4 (the maximum possible). This also affects the explanations for students' scores, harming the explicability of this method.

## 4.2 Reference Aided Evaluation

Reference Aided Evaluation scoring's results are illustrated in Figure 4 (Llama3.1) and Figure 5 (DeepSeek-R1). The variety of scores is much greater than the previous method, which suggests that the given evaluation guidelines are being taken into consideration by the model. The frequency of scores follows a normal distribution. The results show stricter scoring compared to human evaluation, especially when using DeepSeek-R1.

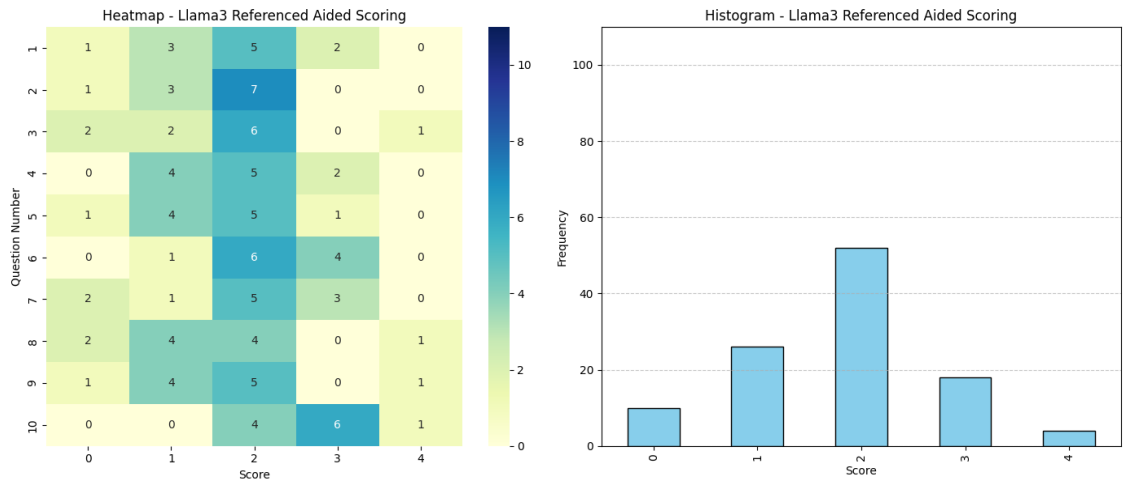

**Figure 4. Heatmap and histogram of scoring results obtained by Llama-3.1 through Reference Aided Evaluation.**

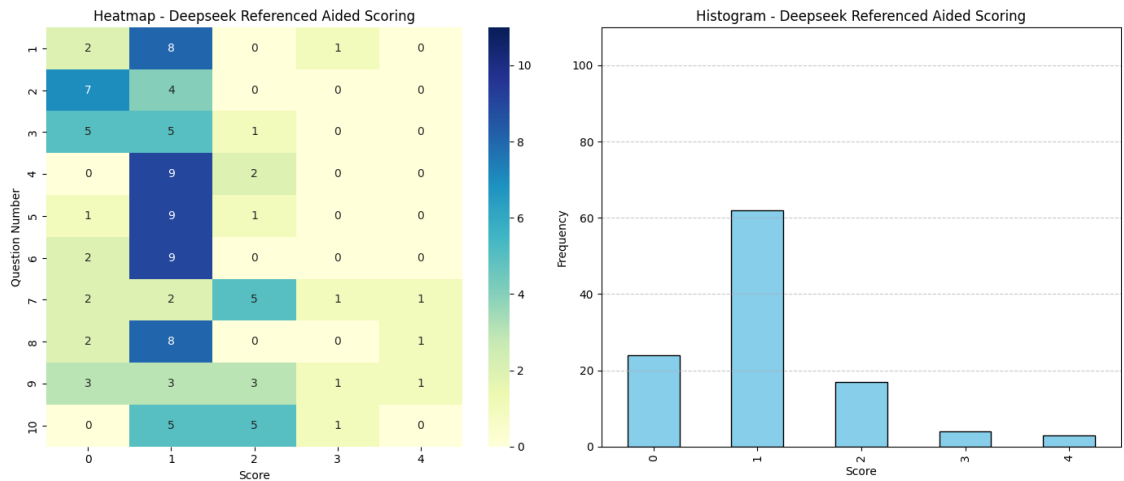

**Figure 5. Heatmap and histogram of scoring results obtained by DeepSeek-R1 through Reference Aided Evaluation.**

Out of the tested methods, Reference Aided Evaluation offers the best explanation and analysis in answers. To show the capacities and drawbacks of explanations produced by this method, two examples are analyzed: one with adequate evaluations and another one with faulty ones.

**Table 4. Reference Aided Evaluation correct example.**

| **Reference Aided Evaluation** (adequate example) | |
|---|---|
| Question | What is the main idea behind object-oriented programming? |
| Reference Answer | Object-oriented programming allows us to bundle together objects that share common properties and procedures, creating an abstraction that lets us distinguish between how to implement an object and how to use it. |
| Student | To encapsulate data and behavior. The main objective is to create a clean, organized and maintainable code, abstracting all the complexity inside the objects. |
| Human Evaluator | Score: 4<br>Talks about the key point of abstraction and offers examples of the advantages of using this programming methodology. Overall a complete and clear answer to the question. |

| AI Evaluator (Llama) | Score: 4<br>The student's answer shows a thorough understanding of the concept of object-oriented programming. They correctly identify the main idea behind object-oriented programming as encapsulating data and behavior, and creating a clean, organized, and maintainable code. They also mention abstracting all the complexity inside the objects, which is a key aspect of object-oriented programming. The student's answer is clear, concise, and accurately reflects the concept. The only thing that could make it even better is if the student provided more specific examples or details to support their answer, but overall, their answer is excellent. |
|---|---|
| AI Evaluator (DeepSeek) | Score: 4<br>The student's answer demonstrates a thorough understanding of object-oriented programming concepts. They correctly identify the main idea as encapsulation of data and behavior, and they elaborate on how this leads to clean, organized, and maintainable code by abstracting complexity. The student's explanation aligns well with the principles of object-oriented programming, showing a deep grasp of the concept. |

For the evaluation shown in Table 4, both Llama3.1 and DeepSeek-R1 award the same score as our human evaluator. Furthermore, the explanations correctly assess the score, and criteria followed, offering deeper insights into the remarkable elements of the student's answer.

Both correctly correlate the concepts between the students' answer and the reference answer despite the differences. With the right techniques, LLMs show promising reasoning capabilities in grading answers (Kortemeyer, 2023). In the examples tested, Llama3.1 tends to provide longer and more insightful feedback than DeepSeek-R1. This could be due to the omission of the thought process of the DeepSeekR1 model in the output, meaning that we lose some of the information of judgement in said thought process.

**Table 5. Reference Aided Evaluation faulty example.**

| **Reference Aided Evaluation** (faulty example) | |
|---|---|
| Question | What is the main goal of analyzing algorithms to determine their complexity? |
| Reference Answer | The main goal is to understand the characteristics of an algorithm so that when creating one's own algorithm, they have a sense of how difficult it will be to solve the problem. |
| Student | The main idea behind this analysis is to know and understand the efficiency of the algorithm. Given that the complexity will tell us information about its speed, memory consumption, etc. In this way, it is important to know how complex the algorithm is and if it is possible to enhance it. |
| Human Evaluator | Score: 4<br>Talks about resource consumption, speed and comparison of algorithms to enhance their performance. Touches up on all key topics and offers examples, getting a score of 4. |
| AI Evaluator (Llama) | Score: 3<br>The student's answer shows a partial understanding of the relevant knowledge. They correctly identify that the main goal of analyzing algorithms to determine their complexity is to understand the efficiency of the algorithm, which includes its speed and memory consumption. However, the answer is not complete as it does not mention the importance of understanding the characteristics of an algorithm to predict its difficulty in solving a problem, which is mentioned in the reference answer. Additionally, the student's answer does not provide any specific examples or details about the characteristics of quadratic complexity algorithms, which are discussed in the context. Overall, |
| | the student's answer demonstrates a good understanding of the general concept, but lacks depth and specificity. |
| AI Evaluator (DeepSeek) | Score: 2<br>The student's answer shows partial understanding of the relevant knowledge. The answer is not complete or contains wrong information. |

Now, in the example in Table 5, each model suffers from a different fault. Llama3.1's evaluation shows clear over-reliance on the reference answer. Predicting difficulty is a topic explicitly stated in the reference answer while the student talks about the importance of knowing the complexity of the algorithm. Furthermore, the model pays close attention to the base material, which is about quadratic complexity algorithms and expects this topic to be reflected on the answers when the question does not demand it as an example. In text-input problems, concepts might be asked to be applied to wider areas of knowledge or with other examples, like in this case. The model cannot properly recognize that the example offered in the lesson content is not relevant to the question.

While this over-reliance on the provided content happens to both models, DeepSeek-R1 showcases a unique problem that is not present in the Llama-3.1's testing. Sometimes the model forgoes explaining its reasoning for the assigned score. The only explanation provided is a copy of the directives offered on the prompt and not any further comments about the student's answers. This lack of consistency makes the fairness of the evaluations unreliable.

## 4.3 No Reference Evaluation

No Reference Evaluation uses the same prompt as Reference Aided Evaluation but without the inclusion of a reference answer. Therefore, only the differences between them are highlighted in this section. The inclusion of this method offers some insights regarding the possible benefits versus the drawbacks of using a reference answer to generate an evaluation. As shown in Figure 6 and Figure 7, the scores generated by this method are lower but retain the same variety as its counterpart evaluation with reference. This differs from the high results obtained by human evaluation.

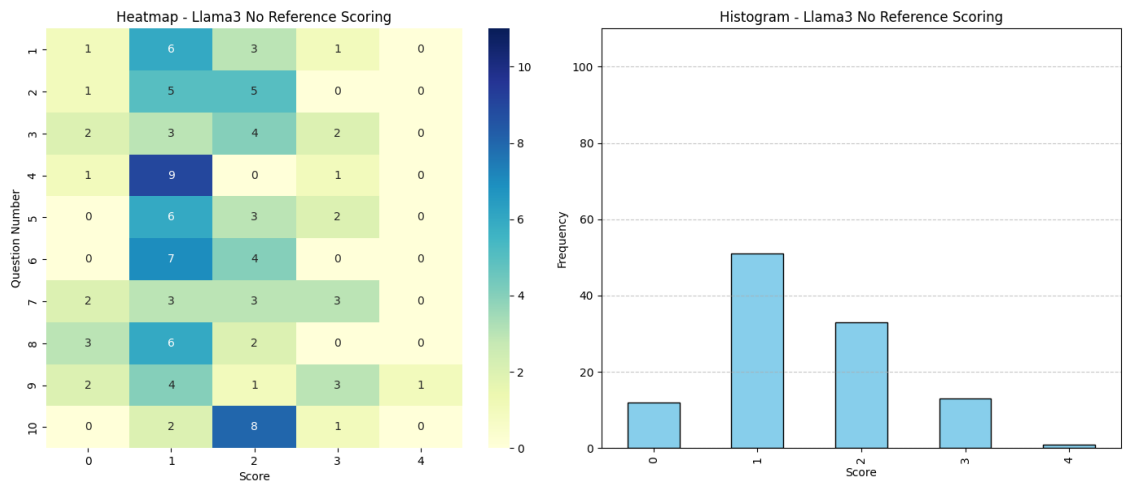

**Figure 6. Heatmap and histogram of scoring results obtained by Llama3.1 through No Reference Evaluation.**

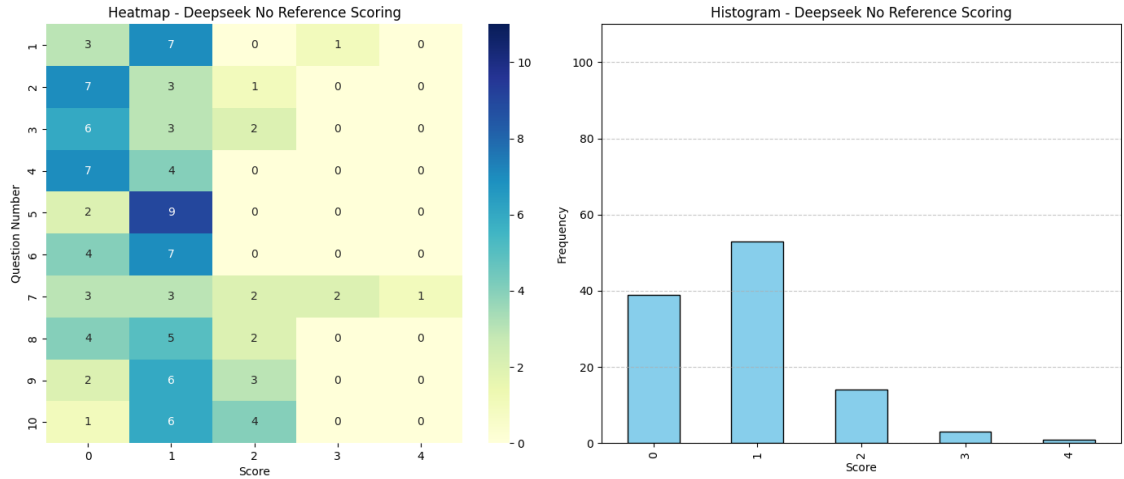

**Figure 7. Heatmap and histogram of scoring results obtained by DeepSeek-R1 through No Reference Evaluation.**

While text-input problems allow for multiple correct answers, they also have the intent to test the student's knowledge, making them draw conclusions and analysis that might not be included in the educational materials. By having these materials as sole context, the model might not be capable of finding the answer to the question. For example, a question about offering examples might be related to a text that explains a concept but does not offer those examples, forcing the student to relate the concept with things they previously knew.

**Table 6. No Reference Evaluation faulty example.**

| No Reference Evaluation | |
|---|---|
| Question | What is the complexity of a function h(n) that adds up the digits of a number together? |
| Reference Answer | The complexity of the function h(n) is logarithmic in the size of the input n, because it reduces the size of the problem by a constant factor on each stage. |
| Student | O(log10(n)) where n is the number and log10(n) the number of digits of n. |
| Human Evaluator | Score: 4<br>The answer is correct and properly relates the complexity with the "n" that has been set in the question. |
| AI Evaluator (Llama) | Score: 0<br>The student's answer is nonsensical and unrelated to the question. The question is asking about the complexity of a function that adds up the digits of a number together, but the student's answer is discussing the complexity of a function that finds the cube root of a number. The student's answer also mentions log10(n) as the number of digits of n, which is not relevant to the complexity of the function h(n) that adds up the digits of a number together. |
| AI Evaluator (DeepSeek) | Score: 1<br>The student's answer shows a partial understanding of the concept.<br>They correctly identified that the complexity is related to the number of digits, which is log10(n). However, their answer lacks the correct reasoning and does not fully explain the reasoning behind the complexity. The correct reasoning involves the number of digits, which is log10(n), and the number of operations needed to sum the digits, which is proportional to the number of digits. Therefore, the complexity is O(log10(n)). The student's answer is partially correct but does not provide a complete or nuanced analysis. |

As shown in Table 6, No Reference Evaluation presents clear problems when facing a mathematical or complex problem. In the example, the student's answer is correct, but the Llama model thinks the actual answer for the question is different. Since no reference answer is provided, the model itself should be able to answer the question correctly in order to evaluate the answer properly. This can be challenging for questions that rely on specific information or mathematical knowledge, as shown in this example. Due to the importance of accuracy when evaluating, this presents a great drawback on the application of this approach.

## 4.4 Additive Evaluation

In regards to additive evaluation, Figure 8 and Figure 9 show the obtained results lack the positive diversity observed in our previous evaluations, most likely due to the restrictive nature of this technique. When generalizing evaluation criteria, this strength becomes a problem. Whilst in the previous methods the interpretation capabilities of LLMs played in our favor, this approach limits the model's judgement, forcing it to perform a checklist of pre-filled criteria.

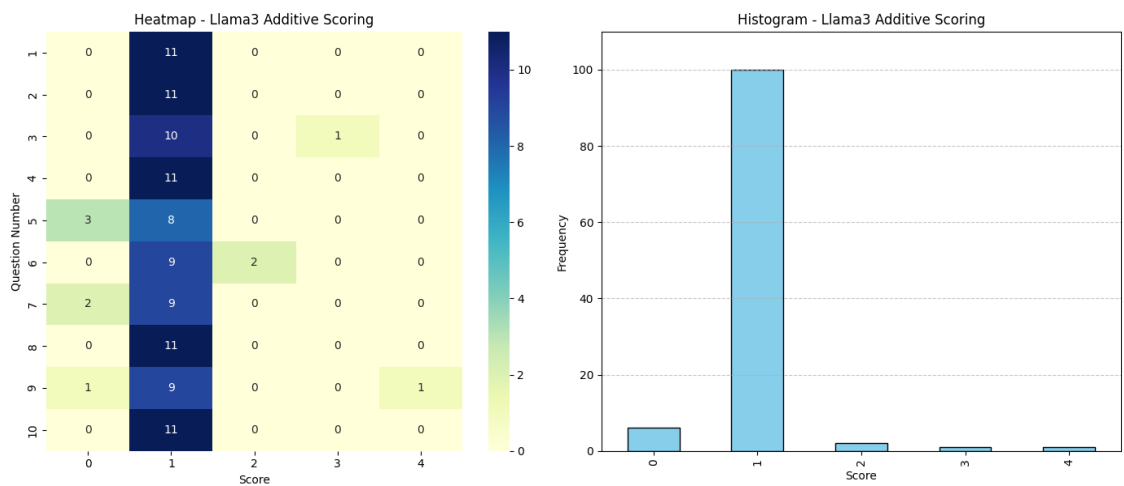

**Figure 8. Heatmap and histogram of scoring results obtained by Llama3.1 through Additive Evaluation.**

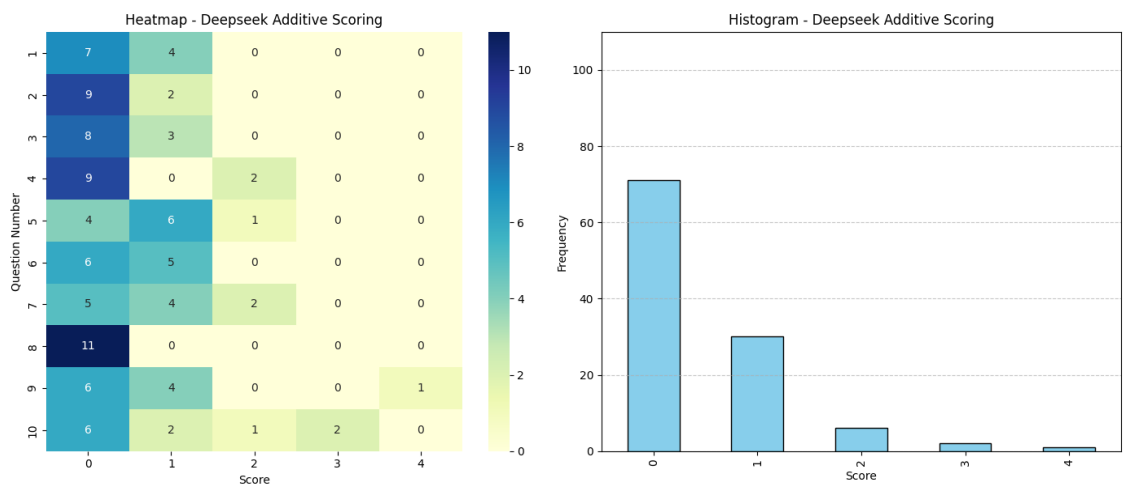

**Figure 9. Heatmap and histogram of scoring results obtained by DeepSeek-R1 through Additive Evaluation.**

As explained previously, the total score of an answer (0-4) in Additive Evaluation is not assigned but rather calculated by adding the points obtained for every criterion an answer meets. The three criteria are correctness (2 points), clarity (1 point) and proper explanation (1 point). To analyze the cause of the poor results obtained with this approach, Table 7 and Table 8 have been included to show the criterion satisfaction rate.

A relevant measure is the incredibly low percentage of answers that are marked as correct (Criteria 1): only 3.64% and 5.45% for Llama3.1 and DeepSeek-R1, respectively. This score does not align with our human evaluation. To compare with our baseline (human evaluation), we take the consideration that a human score of 3 or 4 indicates the answer is at least correct and clear, with only minimal mistakes. By this consideration, 57.27% of answers (63 out of 110) would be considered correct. With a stricter standard, considering only answers with a perfect score of 4 as correct, 29.09% (32 out of 110) would meet the criterion. Neither of these percentages align (or even come close) with those obtained through the automated LLM grading.

While Llama3.1 finds most of the answers clear (Criteria 2) and almost a quarter of them contain correct explanations and reasoning (Criteria 3), DeepSeek-R1's results with this method follow a much more binary pattern (i.e., all criteria are scored as correct or incorrect), where all criteria are correct the same number of times. To analyze the reason behind these results, Table 9 offers an in depth look on the actual values of the results.

Table 7. Frequencies for criteria fulfilment obtained by Llama3.1 for Additive Evaluation.

| Additive Scoring Llama3.1 | Number of Answers that meet the Criteria | Percentage of Answers that meet the Criteria |
|---|---|---|
| Criteria 1 (Correctness) | 4 | 3.64% |
| Criteria 2 (Clarity) | 76 | 69.09% |
| Criteria 3 (Well-Explained) | 27 | 24.55% |

Table 8. Frequencies for criteria fulfilment obtained by DeepSeek-R1 for Additive Evaluation.

| Additive Scoring DeepSeek-R1 | Number of Answers that meet the Criteria | Percentage of Answers that meet the Criteria |
|---|---|---|
| Criteria 1 (Correctness) | 6 | 5.45% |
| Criteria 2 (Clarity) | 6 | 5.45% |
| Criteria 3 (Well-Explained) | 6 | 5.45% |

Table 9. Combination of correct and incorrect criteria of all answers of DeepSeek-R1's Additive Evaluation. All other combinations not shown in the table have zero answers.

| Criterion 1 | Criterion 2 | Criterion 3 | Number of answers |
|---|---|---|---|
| False | False | False | 103 |
| True | True | True | 5 |
| True | True | False | 1 |
| False | False | True | 1 |

As shown in Table 9, 103 of the total answers were evaluated as completely wrong in all criteria, while 7 answers fulfill at least one of them. However, from these 7 answers, 5 of them are completely correct. This means the model is not interpreting each of the criteria independently and instead is trying to look for right and wrong answers (assigning completely correct criteria for answers considered correct), thus creating a binary output where 98.18% of answers are either entirely correct or incorrect, contrary to the expectations of using atomic criteria.

In terms of the evaluations generated, the biggest concern with this approach is inconsistency. As shown in Table 10, particularly in Llama3.1's evaluation, the explanation indicates that the student has a good understanding of the concept, and the answer is accurate. Still, criteria for correctness and clearness are marked as false. As it stands, this method is unreliable to produce fair evaluations.

Table 10. Additive Evaluation faulty example.

| Additive Evaluation | | |
|---|---|---|
| Question | What is the complexity of a function h(n) that adds up the digits of a number together? | |
| Reference Answer | The complexity of the function h(n) is logarithmic in the size of the input n, because it reduces the size of the problem by a constant factor on each stage. | |
| Student | O(log10(n)) where n is the number and log10(n) the number of digits of n. | |
| Human Evaluator | Score: 4<br>The answer is correct and properly relates the complexity with the "n" that has been set in the question. | |
| AI Evaluator (Llama) | Correct? (2 points) No | The student's answer shows a complete and correct understanding of the concept. The function h(n) that adds up the digits of a number together indeed has a time complexity of O(log10(n)) because it needs to iterate over the digits of the number, and the number of digits is proportional to the logarithm of the number. The student correctly identifies the number of digits as the relevant factor in determining the complexity of the function. |
| | Clear? (1 point) No | |
| | Explained? (1 point) Yes | |
| AI Evaluator (DeepSeek) | Correct? (2 points) No | The student's answer is nonsensical or unclear. It is not related to the question. |
| | Clear? (1 point) No | |
| | Explained? (1 point) No | |

## 4.5 Adaptive Evaluation

Results obtained by adaptive evaluation (Figure 10 and Figure 11) are similar to those of No Reference Evaluation for both Llama3.1 and DeepSeek-R1 models. While this method does not possess the weaknesses of No Reference Evaluation (lack of knowledge about a proper answer to the question), the creation of custom-made criteria for short text-input problems worsens the over-reliance on the reference answer. Because these criteria are derived directly from the reference answer, they inherently encode the structure, examples, and level of detail present in that answer. This means all aspects of the reference answer are now expected in the students' answer, transforming the reference answer from a guideline to a requirement.

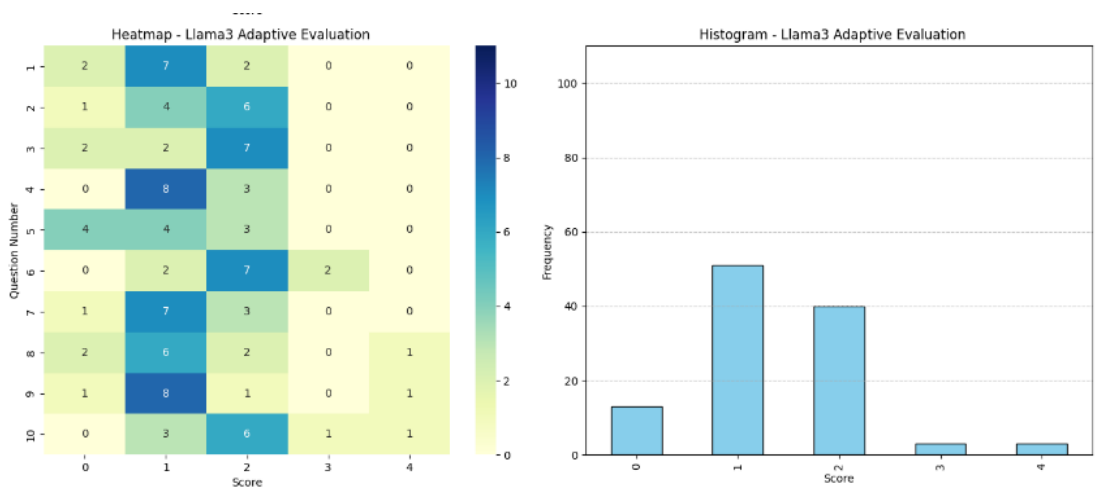

**Figure 10. Heatmap and histogram of scoring results obtained by Llama3.1 through Adaptive Evaluation.**

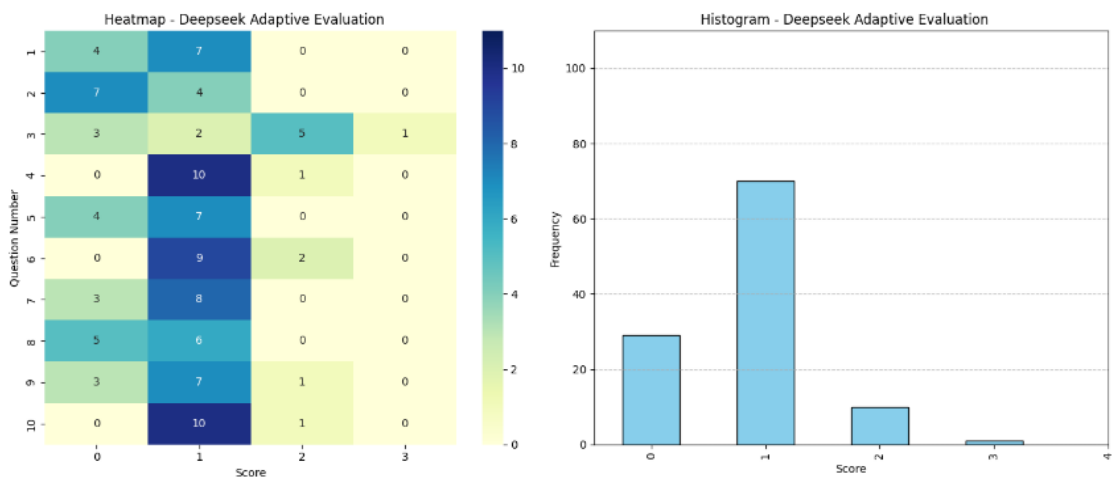

**Figure 11. Heatmap and histogram of scoring results obtained by DeepSeek-R1 through Adaptive Evaluation.**

This over-reliance on a single reference not only narrows the scope of acceptable answers but also undermines the fairness and generality of the evaluation process. Valid answers that approach the problem from different perspectives or use different terminology may be unfairly penalized, even if they demonstrate sound reasoning or factual accuracy. These results do not align with the objective of utilizing text-input problems.

**Table 11. Comparison between evaluators in Reference Aided Evaluation and Adaptive Evaluation.**

| Adaptive Evaluation compared with Reference Aided Evaluation | |
|---|---|
| Question | What is the complexity of a function h(n) that adds up the digits of a number together? |
| Reference Answer | The complexity of the function h(n) is logarithmic in the size of the input n, because it reduces the size of the problem by a constant factor on each stage. |
| Student | O(log10(n)) where n is the number and log10(n) the number of digits of n. |
| Human Evaluator | Score: 4<br>The answer is correct and properly relates the complexity with the "n" that has been set in the question. |

| Llama3.1 | |
|---|---|
| Reference Aided Evaluator | Score: 3<br>The student's answer shows partial understanding of the relevant knowledge. The answer is not complete or contains wrong information. The student mentions the goal of analyzing algorithms to determine their complexity as checking the efficiency of the algorithm, which is partially correct. However, the reference answer highlights that the main goal is to understand the characteristics of an algorithm so that when creating one's own algorithm, they have a sense of how difficult it will be to solve the problem. This suggests that the student's answer lacks the nuance and completeness of the reference answer, scoring a 3. |
| Adaptive Evaluator | Score: 2<br>The student's answer provides a general statement about the importance of understanding algorithm complexity, but does not specifically address the main goal of analysis. It mentions checking the efficiency of the algorithm, which is related to the topic, but lacks detail and clarity. The reference answer provides a more comprehensive explanation of the main goal, which is to understand the characteristics of an algorithm and predict its difficulty in solving a problem. The student's answer does not demonstrate a deep understanding of the topic, but rather a superficial understanding of the importance of efficiency. |
| DeepSeek | |
| Reference Aided Evaluator | Score: 3<br>The student's answer shows a complete and correct understanding of the concept. The answer not only identifies the main goal but also provides a nuanced perspective by mentioning the comparison aspect of efficiency, which adds depth to their understanding. |
| Adaptive Evaluator | Score: 1<br>The student's answer provides a general statement about the importance of understanding algorithm complexity but does not specifically address the main goal of analysis. While it mentions checking efficiency and comparing implementations, it lacks depth and clarity regarding the primary objective of analyzing algorithms, which is to understand their characteristics and predict their difficulty in solving problems. |

This over-reliance on the reference answer can be seen in Table 11. In this example, both Llama 3.1 and DeepSeek-R1 models assign lower ratings to the same student's answer when additional criteria are introduced. Overly strict evaluations are a recurring issue across multiple evaluation methodologies, but they are especially amplified in this adaptive approach.

Given this limitation, adaptive evaluation may be more appropriately applied in contexts where multiple reference answers can be aggregated to create more balanced and inclusive criteria. For example, in essay evaluations or complex problem-solving tasks, multiple high-quality responses could be used to synthesize a more generalized rubric that captures a wider range of acceptable answers.

## 4.6 Comparison with human evaluation

After analyzing the strengths and weaknesses of each individual method, the focus returns to the original baseline: human evaluation. This section presents the deviation, along with the Mean Absolute Deviation (MAD) and Root Mean Square Deviation (RMSE). These data have been calculated for every method except JudgeLM, as the number of generated evaluations in that experimental condition is insufficient for these comparisons (please remember that due to the model's token limitations, 33.63% of answers could not be graded).

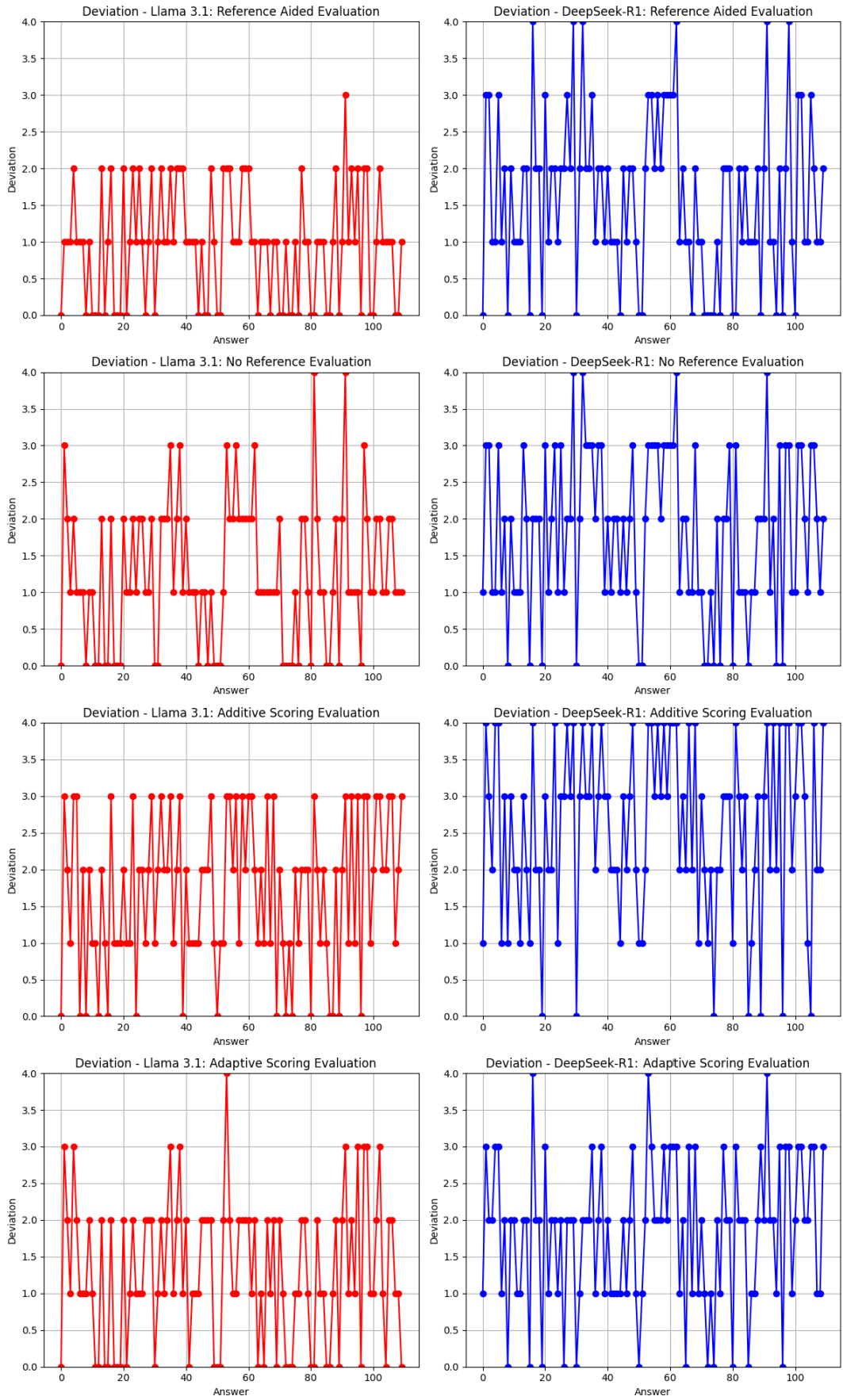

**Figure 12. Deviation from human scoring for every evaluation method for Llama-3.1 and DeepSeek.**

**Table 12. Mean Absolute Deviation (MAD) and Root Mean Square Deviation (RMSE) from human scoring for every evaluation method for Llama-3.1 and DeepSeek. Best results are in bold, second best are underlined.**

|  | Reference Aided Evaluation | No Reference Evaluation | Additive Evaluation | Adaptive Evaluation |
|---|---|---|---|---|
| Llama-3.1-8B | | | | |
| MAD | **0.945** | <u>1.236</u> | 1.682 | 1.273 |
| RMSE | **1.214** | <u>1.555</u> | 1.966 | 1.595 |
| DeepSeek-R1-Distill-Llama-8b | | | | |
| MAD | 1.600 | 1.818 | 2.564 | 1.809 |
| RMSE | 1.944 | 2.119 | 2.828 | 2.065 |

Figure 12 shows the deviation of scores per answer compared to human evaluation for both models. These graphs offer a quick visual representation, allowing us to see that DeepSeek spikes higher than Llama in every evaluation methodology. To further analyze this data, both the MAD and RMSE have been calculated. As seen in Table , Reference Aided Evaluation has the lowest deviation from the human evaluation for both DeepSeek and Llama, being the latter the best overall with the lowest MAD of 0.945 and lowest RMSE of 1.214. This means, on average, the model was one point off from the human evaluator. This suggests that the inclusion of the reference answer helps the model reach similar conclusions to the human evaluator, making it a strong method for generating evaluations.

Other methods, particularly Additive Evaluation, perform worse, reaching a MAD of 1.682 and 2.564 and a RMSE of 1.966 and 2.828 for Llama and DeepSeek respectively. For both models, this is the worst performing methodology, reinforcing the idea that strict atomic criteria limit the model's ability to reason and create open judgements to the presented answer.

## 5 DISCUSSION

From all methodologies and models tested, overall, the Llama3B model had the most satisfactory results as an evaluator, which is in line with previous research in ASAG (Seo et al., 2025). Although our study only evaluated the smaller model, the findings suggest that larger models could yield even better results, as bigger models have proven to have better reasoning and judging capabilities (Thakur et al., 2025). In the following sections, each method's results are explained in depth, as well as their limitations.

## 5.1 Evaluation results for JudgeLM

Overall, JudgeLM's custom approach offers poor results. While the ideas applied for this methodology do offer good results in other approaches (Zhu et al., 2023), adapting a judge from offering feedback for LLM's performance as a chatbot assistant to offer academical assistance seems not possible through prompting. These poor results could be explained by the two following hypotheses.

**Non-customizable grading guidelines.** The model's limited token size impairs its capability to adapt to our needs. This is especially important since the model has been fine tuned to judge LLMs: poor customization means there is a risk of generating poor evaluations to the model's training. While elements such as relevance and clarity remain important factors, others like helpfulness are irrelevant to this purpose. Since JudgeLM is finetuned for evaluating chatbot or assistant LLMs, characteristics like proper vocabulary and politeness are rewarded instead of factuality or reasoning, which does not align with the necessities of academic evaluation. With no way to adapt this through prompting, an untrained JudgeLM model seems unfit for our purposes, as prompt design and instructions affect reliability and performance when obtaining evaluations from LLMs (Chiang & Lee, 2023; P. Wang et al., 2023; Zheng et al., 2023).

**Variable scoring scale**. Since the model has been trained with a scoring scale of 1-10, changing it to 0-4 might worsen the model's understanding of the quality of an answer. While this is a less important drawback than the previous one, ensuring the evaluator model can clearly comprehend scoring scales adds flexibility while maintaining fairness.

Furthermore, explanations for this method proved unfit for our purposes. Since the model has been finetuned to judge other LLMs, without further training and due to max token limitations restricting prompt size and contextualization, JudgeLM cannot provide suitable evaluations for the academic profile. Additionally, JudgeLM's scoring must always be done in pairs, and its single answer scoring is given the reference answer as an answer with perfect score. This means that the model considers the reference answer as another answer to be evaluated, despite already having a score. Therefore, the resulting evaluation offers insight into both the student's answer and the reference answer, adding useless information to its already poor answers.

## 5.2 Evaluation Results for Reference Aided Evaluation

As shown in Table Table 2, LLM evaluation can come close to human evaluation, as also concluded previous studies (Chiang & Lee, 2023). Reference Aided Evaluation was the approach that produced the closest results to human evaluation, presenting stricter but fair evaluations. This could happen for the following reasons.

**Score leniency**. This methodology of evaluation requires the model to fit the answer in one of the categories. Although the understanding of the level of knowledge of the student shown by their answer can be subjective, the model still must look for concrete elements specified in the prompt. A human evaluator can change and adapt their criteria by question, producing more forgiving scoring. Various studies have obtained the same results, with LLM giving higher scores on average than their human evaluator counterparts, raising concerns about reliability (Baral et al., 2024; Flodén, 2025).

**Over-reliance on the reference answer.** Due to the nature of text-input questions, there is not a single concrete answer that is considered correct. This is the main motivation behind using LLMs for this task. When including reference answers as a guide, we further specify what we expect as an answer to the question. However, we run into the risk of penalizing great answers that differ from the reference provided by anchoring the model's judgement on the reference answer, reducing flexibility, which is a common issue in LLM judges (del Gobbo et al., 2023; Zheng et al., 2023; Zhu et al., 2023).

**Language barrier**. Students in this study have different degrees of English proficiency. Whilst the human evaluator was more interested in the content of the answers, the language models may pay closer attention to adequate spelling and phrase structure to judge the student's proficiency on the task.

Despite this, and thanks to the reference answer, this method can offer proper explanations to accompany its evaluation as it does not need to put effort into judging the intent or answer to the question. This is one of the techniques derived from LLM-as-a-judge studies that has had great results for judges (Kim et al., 2024). This allows a greater level of explainability and reliability when using it for generating corrections. The method could be further improved with guidelines for academic level, expected length, etcetera.

## 5.3 Evaluation Results for No Reference Evaluation

No Reference Evaluation has shown poor results, lacking capabilities to correctly interpret criteria and make fair judgements. This aligns with our assumption of the necessity of including a reference answer to ensure quality evaluation. In depth, the poor results could be explained for the following reasons:

**Lack of knowledge.** As explained previously, text-input problems might inquire about a topic by asking for information that is not explicitly contained in it. In technical fields, this presents a clear issue: we depend on the model knowing how to solve or reason the problem correctly, which compromises the integrity of the evaluation (Baral et al., 2024; Flodén, 2025; Kortemeyer, 2023). Llama3.1 particularly suffers greatly from this problem while DeepSeek-R1 shows greater capabilities to correctly reason the answer.

**Poor understanding of the question.** Both models, when offered a reference answer, score it with a 4, the maximum score. On the DeepSeek-R1's case, in both scenarios (judging an answer with and without a reference answer) it possesses the correct answer to the problem as its reasoning skills allow it to generate a correct solution to the problem. However, despite the answer matches the correct solution, it is scored poorly. The reference answer also provides further insights into what is expected as an answer. Here, as an evaluator, DeepSeek-R1 expects a more in-depth analysis that is not the intent of the question.

In conclusion, including a reference answer is a key aspect in obtaining fair evaluations. The benefits of having a point of reference for the solution of a question or problem is especially important, as concluded by Kim et al2024) as well. The possible drawbacks related to the limitation of flexibility or creativity remain to be tested for other fields of study,

## 5.4 Evaluation Results for Additive Evaluation

Additive Evaluation reveals several weaknesses in its current structure, primarily due to its rigid treatment of criteria and underuse of the LLMs' language reasoning capabilities. The following issues explain why this evaluation method may lead to suboptimal scoring outcomes:

**Restriction of free model interpretation**. LLMs are fundamentally designed to interpret language holistically, understanding nuance, intent, and meaning beyond strict keyword matching. However, Additive Evaluation breaks this capacity down into isolated binary checks. By requiring a yes/no decision for each criterion without allowing contextual interpretation, the system undercuts the model's innate strength: language reasoning. This leads to situations where a partially correct or insightful answer might be penalized simply because it doesn't match the rigid formulation of a criterion, even though it demonstrates understanding. While guidelines, step-by-step instructions and solid frameworks improve evaluations (Chiang & Lee, 2023; Kim et al., 2024), this method is too restrictive, undermining the advantages of structured approaches.

**Binary criteria enforcement**. The system requires each criterion to be marked as either fulfilled or not, with no allowance for partial correctness or interpretative nuance. This forces the evaluator model to ignore degrees of correctness, and instead treat all deviations equally. For LLMs, this binary decision-making flattens their reasoning process and does not reflect their capacity for gradient judgments or weighing of alternatives. It also introduces brittleness, where a single missing concept or word can negate otherwise valid reasoning.

**Inconsistency in scoring and justification**. One of the most apparent problems is that the explanations provided by the model do not always align with the binary values assigned to criteria, creating discrepancies between scores and feedback, in line with previous research by (Chiang & Lee, 2023; Seo et al., 2025). This happens because the explanations are generated through natural language generation, while the criteria are evaluated separately as atomic booleans. These two processes, while both LLM-driven, do not always converge. For instance, a model might state that a criterion is fulfilled, and explain it correctly, but the boolean flag shows a failure, leading to internal contradictions. This inconsistency undermines trust in the evaluation and reveals a design flaw in how the evaluation is structured.

In conclusion, the atomic criteria are restrictive and generate inconsistencies, and models perform better when given more room to score independently through their own reasoning. While atomic checks provide clear traceability, they interfere with the model's capacity to evaluate holistically. Future evaluation methods should consider hybrid approaches that combine structured expectations with open-ended reasoning, allowing LLMs to judge based on understanding rather than on strict formal compliance.

## 5.5 Evaluation Results for Adaptive Evaluation

Adaptive Evaluation introduces flexibility by allowing the model to generate custom criteria based on the input. While this technique moves away from fixed, atomic checks and leans more on model-driven understanding, it also presents important limitations tied to how the reference answer is used in generating these criteria. The shortcomings of this method could be due to:

**Overreliance on the reference answer**. Although the model dynamically generates evaluation criteria, these are consistently grounded in the content of the reference answer. As a result, the generated criteria inherit the same biases and specificity of the reference itself, making them overly narrow. Student responses that take valid alternative approaches or emphasize different, but still relevant, aspects of the topic are at risk of being judged unfairly. Instead of acting as a baseline for understanding, the reference answer becomes a rigid anchor that constrains how understanding can be expressed.

**Limitation of expected length and detail**. Because the criteria are tightly coupled to what is present (or absent) in the reference answer, they implicitly encode expectations about how detailed or developed the student response should be. If the reference answer is concise or omits elaborations, the resulting evaluation criteria may penalize valid, well-reasoned answers that follow a different path. This not only narrows the space of acceptable responses but can also unfairly penalize the longer or more creative ones. Having only a single reference answer creates a fragile and biased benchmark.

While Adaptive Evaluation is a promising step towards more intelligent assessment, it inherits significant limitations from its dependence on a single reference answer. This technique is expected to perform better in lengthier, open-ended evaluation formats such as essays, where there is room for nuance, structure, and argumentation. To fulfill its potential, it would greatly benefit from the inclusion of multiple reference answers, allowing it to generate more generalized, inclusive criteria that reflect the diversity of acceptable student responses. This would allow models to better leverage their interpretive abilities and reward valid, varied expressions of understanding.

## 6 CONCLUSIONS

Through a collection of 110 answers for text-input problems in the field of computer science from students in tertiary education, we have tested five different methodologies with one judge, JudgeLM, and two other models: Llama3.1 and DeepSeek-R1-Distill-8B. From these two models, Llama3.1's results were more satisfactory for this experiment. While Llama3 is a general-purpose instruction-following model designed to handle a wide range of tasks with broad applicability, DeepSeek-R1 excels in solving problems through its specialized fine-tuning, which emphasizes the use of the CoT technique for enhanced reasoning capabilities, enabling it to breakdown problems in steps. This means approaches that directly utilize this will present better results for the model.

We suspect the main problem behind DeepSeek-R1's underperformance was not the model itself, but the methods applied. Due to the model's reinforced training, it always outputs its thought process to come with an answer, which we discard since it is not part of what we look for. However, in this truncation, we may lose key reasoning for evaluations. Moreover, when not offered a reference answer, DeepSeek-R1 had greater capabilities to solve problems by itself to grade correctly and showed less inconsistencies overall. This model shows promise to work in this context with a method developed to accommodate its thought process, especially in contexts where a reference answer for a problem is not available.

The method evaluated with JudgeLM did not offer remarkable results. As it stands, LLM-as-judges do not have enough flexibility to adapt from judging models to academic evaluation without further fine-tuning. Particularly, due to the JudgeLM's restrictive token limit, further prompting techniques could not be tested. Other models that include rubrics such as Prometheus[4] (Kim et al., 2024) might perform better in this end.

The following two methods, Reference Aided Evaluation and No Reference Evaluation proved that a reference answer had a positive impact on the results. While it can restrain flexibility and creativity in answers, it also offers important key information about the expected answer. A reference answer not only has the correct content but shows what level of depth and analysis is expected. For academic evaluation, we have found this reference answer to be necessary to ensure the model has guidance to evaluate and does not rely on its own knowledge to solve problems. Due to this, Reference Aided Evaluation was the best performing method in our experiment.

The last two methods, Additive Evaluation and Adaptive Evaluation, were assessed to further explore how flexible this pipeline could be for end users. While the customization of rubrics was a good end, specialized atomic rubrics for concise questions such as text-input problems did not prove to be a suitable method. Both would serve better for longer evaluation tools such as essays, problem resolutions, etc. As it stands, we are unsure how well these methodologies would perform with other evaluation items. In the Additive Evaluation's case, it highlights how restrictive criteria limit the understanding models have of the student's answer, worsening results and creating inconsistencies; i.e., models evaluate better when given room to interpret and judge by themselves. Additionally, Adaptive Evaluation particularly would benefit from a wider variety of reference answers for each question to ensure quality criteria.

With the right methods, LLMs show promise to work as evaluators to help in the academic field, both for students and teachers. To conclude, we discuss the limitations of this study, which include the following. Firstly, lack of model size diversity, as all models tested for this experiment were in the 7-8B parameter range. Better performance could be expected from bigger, more powerful models. Secondly, better adaptability for academic level and difficulty. Currently, Reference Aided Evaluation is the only method that can be flexible for different levels and difficulties. However, this must be inferred by the model judging by the reference answer only.

Lastly, our experiment is limited to answers related to computer science and one reference evaluator. Other fields such as mathematics and arts require different degrees of accuracy, factuality, leniency and creativity, which we have not been able to test. Future work for this line of research include testing the proposed methodologies with a wider range of evaluation items (such as long form essays or process-based problems), disciplines and other LLM of bigger sizes. The proposed methodologies can also be expanded by adding multiple reference answers for Reference Aided Evaluations and Adaptative Evaluations to study if it would improve their performance.

**ACKNOWLEDGEMENTS**

The authors thank the support provided by the project "Using language models and chatbots for building virtual assistants in MOOCs" funded by the MISTI Global Seed Funds program.

---

[4] https://huggingface.co/prometheus-eval/prometheus-7b-v2.0

# ANNEX A

Prompts used for JudgeLM Evaluation with JudgeLM-8B. For our purposes, a single answer scoring method has been employed (Zhu et al., 2023).

---

You are a helpful and precise assistant for checking the quality of the answer.

[Question]

{question}

[The Start of Student 1's Answer]

{reference}

[The End of Student 1's Answer]

[The Start of Student 2's Answer]

{answer}

[The End of Student 2's Answer]

[System]

We would like to request your feedback on the performance of two students in response to the question displayed above.

Please rate the factuality, relevance, accuracy, level of details of their responses. Each assistant receives an overall score on a scale of 0 to 4, where a higher score indicates better overall performance. The highest score one can obtain is 4 and the lowest is 0.

Please first output a single line containing only two values indicating the scores for Student 1 and 2, respectively. The two scores are separated by a space. In the subsequent line, please provide a comprehensive explanation of your evaluation, avoiding any potential bias and ensuring that the order in which the responses were presented does not affect your judgment.

[Response]

---

**Figure 13. JudgeLM adapted prompt.**

# ANNEX B

Prompts used for Reference Aided Evaluation with Llama-3.1-8B and DeepSeek-R1-Distill-Llama-8B.

You are a questionnaire assistant. Your task is to generate an evaluation for a student's answer to a TIP (text-input problem) type question. You will have to provide a "score" scoring how well the student answered the relevant question.

The score you need to give will go in a scale of 0-4. The scoring criteria is as follows:

   0. The student's answer is nonsensical or unclear. It is not related to the question.

   1. The student's answer shows serious misconceptions or lack of understanding of the concept. The answer is factually wrong.

   2. The student's answer shows partial understanding of the relevant knowledge. The answer is not complete or contains wrong information.

   3. The student's answer shows a complete and correct understanding of the concept.

   4. The student's answer shows through understanding of the concept that was asked and offers and nuanced analysis thorough reasoning.

To aid in your evaluation, you will be given four things:

  - [Context]:

     It's the content from which the question was extracted.

     Use this as guidance for judging the information in the student's answer.

     Limit the information you use to the context provided: do not judge based on knowledge that is not explicitly provided.

  - [Question]:

     It's the question that was asked to the student.

     The score you give to the student's answer must be in relation of what was asked in the question.

  - [Reference Answer]:

     A provided reference answer to help you judge. This answer has a score of 4.

     This reference is a guide: prioritize judging based on the context.

     A student's answer does not have to follow the reference answer exactly to score a 4.

     Do not use the reference answer's structure to evaluate: judge just based on its content.

  - [Student Answer]:

     This is the answer you will have to judge.

     You must judge this answer based on content alone. Do not judge its structure or grammar.

Your evaluation, meaning your response, will have the following format:

{

   "score": int      # score from 0-4 given to the student's answer

   "evaluation": str   # your rationale for the rating, as a text. Relate your evaluation to the provided scoring criteria.

}

**Figure 14. System prompt for Reference Aided Evaluation.**

```
[Context]
{context}
[Question]
{question}
[Reference Answer]
{reference_answer}
[Student Answer]
{student_answer}
```

**Figure 15. User prompt for Reference Aided Evaluation and Adaptive Evaluation.**

# ANNEX C

Prompts used for No Reference Evaluation with Llama-3.1-8B and DeepSeek-R1-Distill-Llama-8B.

You are a questionnaire assistant. Your task is to generate an evaluation for a student's answer to a TIP (text-input problem) type question. You will have to provide a "score" scoring how well the student answered the relevant question.

The score you need to give will go in a scale of 0-4. The scoring criteria is as follows:

   0. The student's answer is nonsensical or unclear. It is not related to the question.

   1. The student's answer shows serious misconceptions or lack of understanding of the concept. The answer is factually wrong.

   2. The student's answer shows partial understanding of the relevant knowledge. The answer is not complete or contains wrong information.

   3. The student's answer shows a complete and correct understanding of the concept.

   4. The student's answer shows through understanding of the concept that was asked and offers and nuanced analysis thorough reasoning.

To aid in your evaluation, you will be given four things:

   - [Context]:

      It's the content from which the question was extracted.

      Use this as guidance for judging the information in the student's answer.

      Limit the information you use to the context provided: do not judge based on knowledge that is not explicitly provided.

   - [Question]:

      It's the question that was asked to the student.

      The score you give to the student's answer must be in relation of what was asked in the question.

   - [Student Answer]:

      This is the answer you will have to judge.

      You must judge this answer based on content alone. Do not judge its structure or grammar.

Your evaluation, meaning your response, will have the following format:

{

   "score": int      # score from 0-4 given to the student's answer

   "evaluation": str   # your rationale for the rating, as a text. Relate your evaluation to the provided scoring criteria.

}

**Figure 16. System prompt for No Reference Evaluation.**

```
[Context]
{context}
[Question]
{question}
[Student Answer]
{student_answer}
```

**Figure 17. User prompt for No Reference Evaluation and Additive Evaluation.**

# ANNEX D

Prompts used for Additive Evaluation with Llama-3.1-8B and DeepSeek-R1-Distill-Llama-8B. User prompt can be seen in Figure 17Figure 17.

You are a questionnaire assistant. Your task is to generate an evaluation for a student's answer to a TIP (text-input problem) type question. You will have to provide a "score" scoring how well the student answered the relevant question.

The score you need to give will go in a scale of 0-4. Next, you are given some evaluation criteria to score the answer. Each criteria will have a punctuation and a description. If the student's answer matches the criteria you will add the corresponding punctuation to their score, starting at 0.

The criteria are as follows:

   [C1] Points: 2. Description: the student's answer is factually correct.

   [C2] Points: 1. Description: the student's answer is clear and uses vocabulary related to the topic.

   [C3] Points: 1. Description: the student offers explanation or context for their answer.

For example, if a student's answer meets criteria C1 and C3 their score will be the sum of the points each criteria adds: 2 + 1 = 3. In this example, the student's score is 3. If the student meets none of the criteria, their score will be 0.

To aid in your evaluation, you will be given four things:

   - [Context]:

       It's the content from which the question was extracted.

       Use this as guidance for judging the information in the student's answer.

       Limit the information you use to the context provided: do not judge based on knowledge that is not explicitly provided.

   - [Question]:

       It's the question that was asked to the student.

       The score you give to the student's answer must be in relation of what was asked in the question.

   - [Student Answer]:

       This is the answer you will have to judge.

       You must judge this answer based on content alone. Do not judge it's structure or grammar.

Your evaluation, meaning your response, will have the following format:

{

  "c1": boolean      # does the student meet criteria C1?

  "c2": boolean      # does the student meet criteria C2?

  "c3": boolean      # does the student meet criteria C3?

  "score": int      # score from 0-4 given to the student's answer

  "evaluation": str   # your rationale for the rating, as a text. Relate your evaluation to the provided scoring criteria.

}

**Figure 18. System prompt for Additive Evaluation.**

# ANNEX E

Prompts used for Adaptive Evaluation with Llama-3.1-8B and DeepSeek-R1-Distill-Llama-8B. User prompt can be seen in Figure 15Figure 15.

You are an academic advisor. Your job is to create evaluation criteria to guide the evaluation of text-input problems.

These text-input problems aim to measure the understanding of students about the topic, testing areas such as their knowledge, comprehension, application, analysis, synthesis and evaluation.

The more related evaluation criteria are to the specific topic and question presented, the better. To help you come up with said evaluation criteria you will be offered the following things:

   - [Context]: It's the content from which the question was extracted. Use this to guide your knowledge and ensure your evaluation criteria are adequate and relevant.

   - [Question]: This is the txt-input problem. Your evaluation criteria will revolve around this question. Remember, the job of the evaluation criteria is to measure the student's performance when answering the question.

   - [Reference Answer]: This reference answers the provided question. Use this as a reference to see what could be expected from the student. Be careful relying on it, text-input problems can be answered in many different ways, use this as a guide to identify key topics. A student's answer does not need to match the reference answer to obtain the highest score.

Here are some additional guidelines to make these criteria:

   - Before thinking of the criteria, make sure you understand the question. Analyze these elements and then make evaluation criteria accordingly:

      - What does this question ask for?

      - What is the aim of this question? Does it test reasoning, knowledge, application...?

      - What skills should a student demonstrate to answer the question correctly?

   - Evaluation criteria will be a list of criteria that grants a score from 0-4. Zero is the lowest possible score, answers that score a zero are non-sensical and un-related to the topic. Four is the highest score, answers that score a four must be correct and complete in accordance to the question.

   - Be clear and specific: state what you expect in each scoring tier.

   - Text-input problems award analysis, good answers should be complete and offer good insight, explanations or examples. Only ask for this if it is appropriate for the question: this would not apply for concrete problems that ask for values, for example.

   - Leave room for improvement: students should be rewarded for offering more information as long as it is related to the question. Good explanations show through understanding.

You will write the evaluation criteria as a numbered list. Write only this list, do not include further explanations. Make sure to include all five criterion. Number them 0, 1, 2, 3, and 4.

0. Evaluation criteria expected to award a question a score of 0.

1. Evaluation criteria expected to award a question a score of 1.

2. Evaluation criteria expected to award a question a score of 2.

3. Evaluation criteria expected to award a question a score of 3.

4. Evaluation criteria expected to award a question a score of 4.

**Figure 19. System prompt for Criteria Generation.**

> You are a questionnaire assistant. Your task is to generate an evaluation for a student's answer to a TIP (text-input problem) type question. You will have to provide a "score" scoring how well the student answered the relevant question.
>
> The score you need to give will go in a scale of 0-4. The scoring criteria is as follows:
>
> {criteria}
>
> To aid in your evaluation, you will be given four things:
>
>   - [Context]:
>
>     It's the content from which the question was extracted.
>
>     Use this as guidance for judging the information in the student's answer.
>
>     Limit the information you use to the context provided: do not judge based on knowledge that is not explicitly provided.
>
>   - [Question]:
>
>     It's the question that was asked to the student.
>
>     The score you give to the student's answer must be in relation of what was asked in the question.
>
>   - [Reference Answer]:
>
>     A provided reference answer to help you judge. This answer has a score of 4.
>
>     This reference is a guide: prioritize judging based on the context.
>
>     A student's answer does not have to follow the reference answer exactly to score a 4.
>
>     Do not use the reference answer's structure to evaluate: judge just based on its content.
>
>   - [Student Answer]:
>
>     This is the answer you will have to judge.
>
>     You must judge this answer based on content alone. Do not judge its structure or grammar.
>
> Your evaluation, meaning your response, will have the following format:
>
> {
>
>   "score": int      # score from 0-4 given to the student's answer
>
>   "evaluation": str   # your rationale for the rating, as a text. Relate your evaluation to the provided scoring criteria.
>
> }

**Figure 20. System prompt for Adaptive Evaluation.**

Here is the [Context], [Question] and [Reference Answer]. Generate evaluation criteria for the presented scenario:

[Context]

{context}

[Question]

{question}

[Reference Answer]

{reference_answer}

**Figure 21. User prompt for Criteria Generation.**